\documentclass[letterpaper, 10 pt, conference]{ieeeconf}  

\IEEEoverridecommandlockouts                              

\usepackage{times} 
\usepackage{subcaption}
\usepackage{pifont}
\usepackage{amsmath} 
\usepackage{amssymb}  
\newcommand{\cmark}{\ding{51}}%
\newcommand{\xmark}{\ding{55}}%

\usepackage{multirow}
\usepackage{graphicx}
\usepackage{booktabs}
\usepackage{orcidlink}
\usepackage{makecell}
\usepackage{subcaption}
\usepackage{hyperref}
\hypersetup{
    colorlinks=true,
    urlcolor=blue
}

\newcommand{\rev}[1]{#1}

\title{\LARGE \bf
FALCON: Future-Aware Learning with Contextual Object-Centric Pretraining for UAV Action Recognition
}

\author{
Ruiqi Xian$^{1}$, Xiyang Wu$^{1}$, Tianrui Guan$^{1}$, Xijun Wang$^{1}$, Boqing Gong$^{2}$ and Dinesh Manocha$^{1}$
}

\begin{document}

\maketitle
\thispagestyle{empty}
\pagestyle{empty}

\begin{abstract}
We introduce FALCON, a unified self-supervised video pretraining approach for UAV action recognition from raw RGB aerial footage, requiring no additional preprocessing at inference.
UAV videos exhibit severe spatial imbalance: large, cluttered backgrounds dominate the field of view, causing reconstruction-based pretraining to waste capacity on uninformative regions and under-learn action-relevant human/object cues.
FALCON addresses this by integrating object-aware masked autoencoding with object-centric dual-horizon future reconstruction.
Using detections only during pretraining, we construct objectness priors that (i) enforce balanced token visibility during masking and (ii) concentrate reconstruction supervision on action-relevant regions, preventing learning from being dominated by background appearance.
To promote temporal dynamics learning, we further reconstruct short- and long-horizon future content within an object-centric supervision region, injecting anticipatory temporal supervision that is robust to noisy aerial context.
Across UAV benchmarks, FALCON improves top-1 accuracy by 2.9\% on NEC-Drone and 5.8\% on UAV-Human with a ViT-B backbone, while achieving 2$\times$--5$\times$ faster inference than supervised approaches that rely on heavy test-time augmentation.
\end{abstract}
\section{Introduction}
\label{introduction}

Unmanned aerial vehicles (UAVs) equipped with onboard cameras enable a wide range of vision-based robotics applications, including human detection~\cite{gong2020effective}, tracking~\cite{wang2019real}, and surveillance~\cite{zhu2021detection}.
Among these, action recognition from aerial views is particularly important for high-level perception tasks such as search-and-rescue, intent estimation, and human--robot collaboration, where understanding human motion supports timely and autonomous decision-making.

\begin{figure}[tp]
    \centering
    \includegraphics[width=0.75\columnwidth]{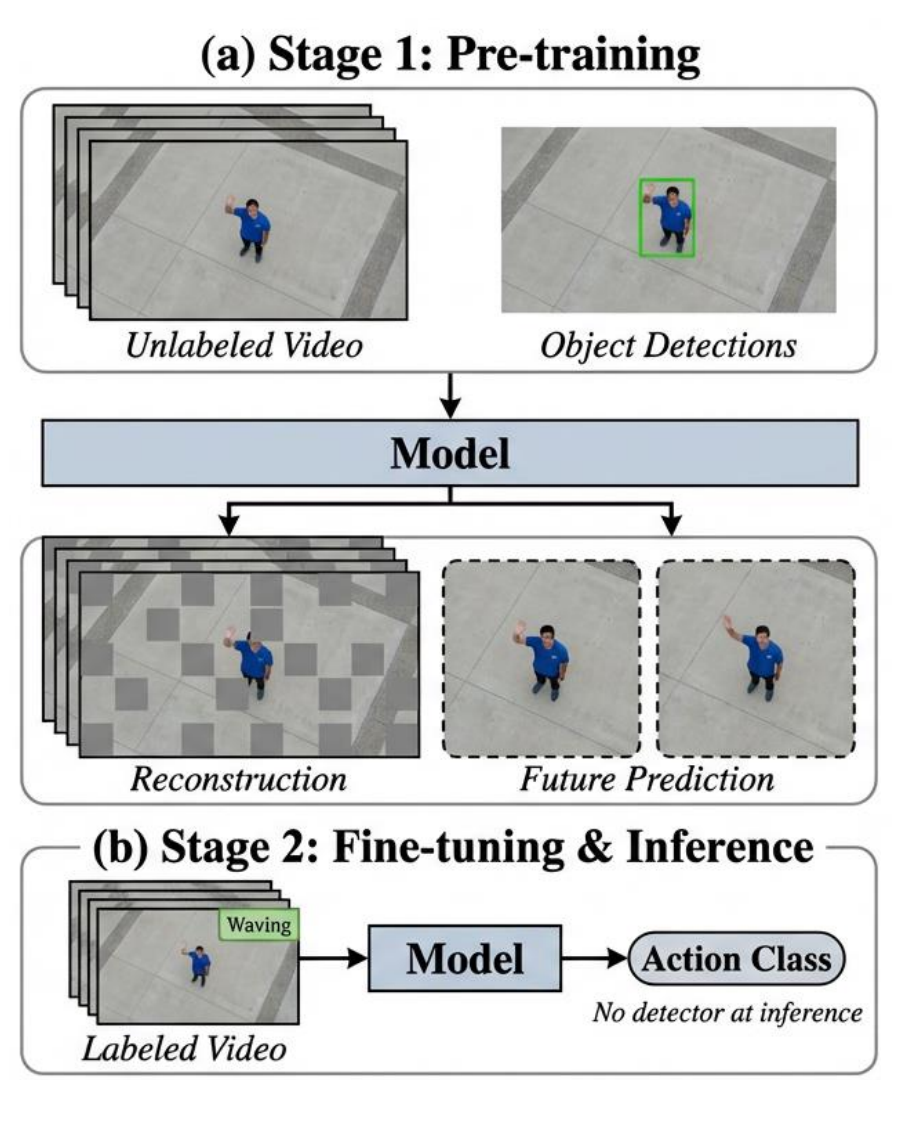}
    \vspace{-10pt}
\caption{\textbf{Overview of FALCON.}
\textbf{Stage 1 (Pre-training):} Given unlabeled UAV videos, we use off-the-shelf detections \emph{only during pretraining} to instantiate objectness cues and optimize an object-aware masked reconstruction objective together with object-centric dual-horizon future reconstruction.
\textbf{Stage 2 (Fine-tuning \& Inference):} The pretrained model is transferred to UAV action recognition using labeled videos, and inference is performed end-to-end from raw RGB without detectors or region processing.}
\label{fig:cover}

    \label{fig:cover} 
\vspace{-20pt}
\end{figure}

Despite recent progress, UAV-based action recognition remains challenging.
Compared to ground-view videos, UAV footage exhibits a distinctive visual structure: action-relevant humans/objects are often tiny, while large cluttered backgrounds and camera-induced motion dominate the scene.
This makes fine-grained motion understanding difficult and encourages models to exploit spurious background correlations.
Meanwhile, labeled aerial datasets are limited and expensive to curate---for example, UAV-Human~\cite{li2021uav} contains only 22k videos, far smaller than large-scale ground-view datasets such as Kinetics~\cite{kay2017kinetics}---which further motivates learning from unlabeled aerial video.

Self-supervised learning (SSL) is therefore a natural direction for UAV action recognition.
However, UAV videos expose a fundamental mismatch for standard reconstruction-based objectives.
Masked autoencoding (MAE)~\cite{feichtenhofer2022masked} learns representations by reconstructing masked tokens from visible context, yet under extreme background dominance, random masking and uniform reconstruction allow the learning signal to be explained primarily by background textures.
As a result, small human/object regions are frequently missing from the visible set and contribute weakly to optimization, yielding representations that under-emphasize action-carrying motion cues.

Beyond the spatial long-tail, aerial action recognition also demands learning motion evolution.
Conventional masked reconstruction objectives mainly supervise recovering missing content within the observed segment, which can often be satisfied by modeling local appearance and short-range spatiotemporal smoothness.
For UAV videos, where action cues are subtle and targets are small, such supervision provides limited pressure to encode how human/object regions evolve over time.

A natural remedy is to incorporate future modeling.
However, naively reconstructing full future frames in aerial footage is easily dominated by ego-motion and background clutter, again allocating capacity to regions weakly related to actions.
This suggests that effective aerial SSL should re-target both spatial supervision and temporal objectives toward action-carrying regions, and explicitly capture motion evolution across multiple horizons.

\textbf{Main Contribution:} Motivated by these observations, we propose \textbf{FALCON}, a UAV-tailored self-supervised pretraining objective that explicitly centers learning on action-carrying human/object regions and their temporal evolution (Fig.~\ref{fig:cover}).
FALCON uses off-the-shelf object cues only during pretraining to mitigate background-dominated learning, and introduces a future-aware objective to encourage anticipatory motion understanding.
Importantly, FALCON requires no detector input, bounding boxes, or extra test-time augmentation during fine-tuning or inference. The main contributions of our work include:
\begin{itemize}
    \item \textbf{UAV-specific objective diagnosis.} We identify two objective mismatches in UAV SSL---background-dominated reconstruction learning and ego-motion/background contamination in temporal objectives---that make standard reconstruction-based pretraining poorly aligned with action semantics.
    \item \textbf{Object-aware masked modeling for UAV videos.} We introduce an object-aware masked pretraining formulation that jointly designs token visibility and reconstruction supervision around action-carrying regions, countering the extreme background dominance in aerial footage.
    \item \textbf{Object-centric dual-horizon temporal objective.} We introduce an object-centric dual-horizon future reconstruction objective that targets short- and long-horizon motion evolution within an object-focused supervision region, providing anticipatory supervision with no inference overhead.
\end{itemize}
In practice, FALCON improves action recognition on UAV benchmarks.
With a ViT-B backbone, it improves top-1 accuracy by 2.9\% on NEC-Drone and 5.8\% on UAV-Human, while delivering 2$\times$--5$\times$ faster inference than supervised approaches that rely on heavy test-time augmentation.
\section{Related Work}
\label{relatedwork}

\begin{figure*}[!h]
    \centering
    \includegraphics[width=0.8\textwidth]{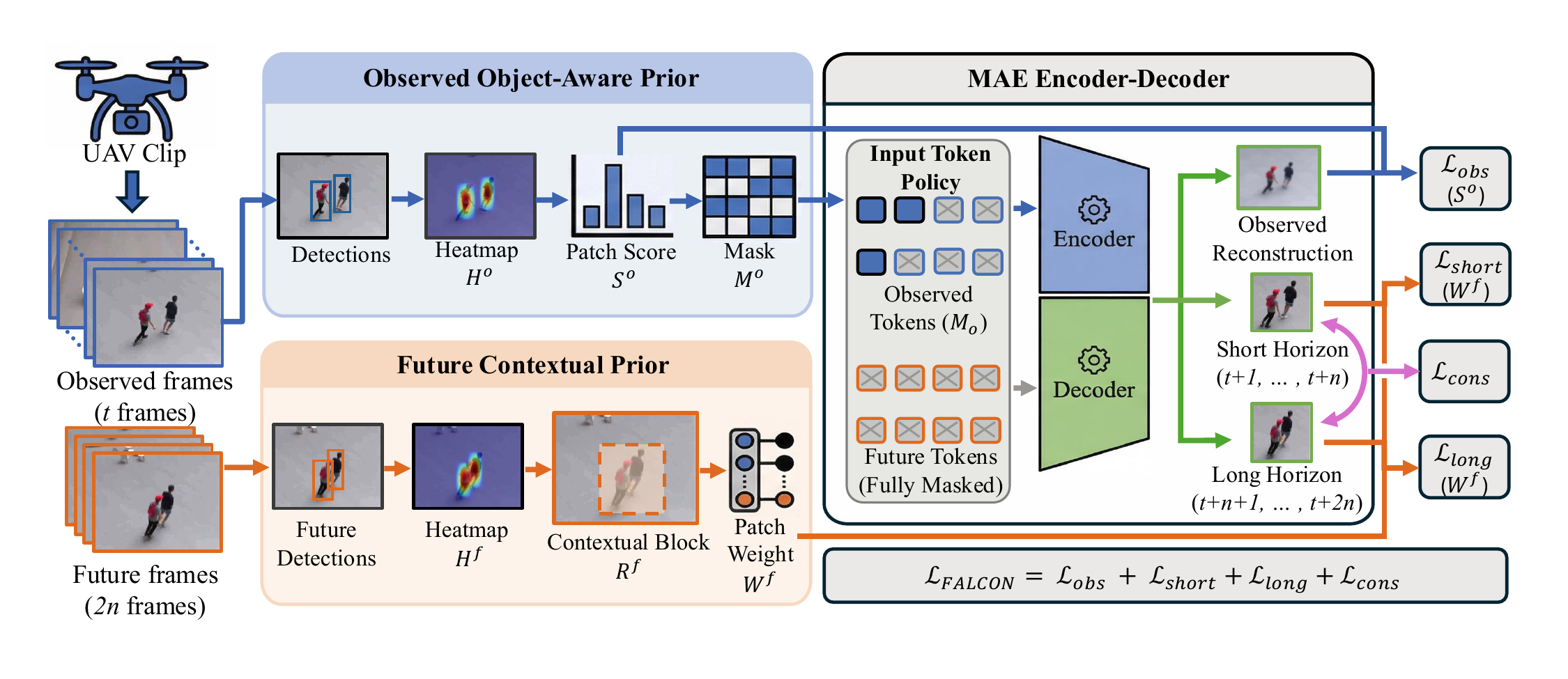}
    \vspace{-10pt}
\caption{\textbf{FALCON pipeline.}
Given a UAV clip, we split it into observed and future frames.
Pretraining-time detections instantiate an \emph{observed object-aware prior} ($H^o\!\rightarrow\!S^o\!\rightarrow\!M^o$) to enable stratified masking and object-centric supervision for observed reconstruction, and a \emph{future contextual prior} ($H^f\!\rightarrow\!\mathcal{R}^f\!\rightarrow\!W^f$) that restricts \emph{where} dual-horizon future reconstruction is supervised.
An MAE encoder--decoder reconstructs observed tokens and fully masked future tokens with short/long losses and a horizon-consistency regularizer, optimized by $\mathcal{L}_{\textsc{falcon}}=\mathcal{L}_{obs}+\mathcal{L}_{short}+\mathcal{L}_{long}+\mathcal{L}_{cons}$.}

\label{fig:pipeline}
    \label{fig:pipeline} 
\vspace{-15pt}
\end{figure*}

\textbf{Action Recognition for UAV Videos.}
Deep learning has significantly advanced action recognition in ground-based videos~\cite{soans2020sa,massardi2020parc,yao2022pa}, yet UAV scenarios remain particularly challenging due to strong camera motion, diverse viewpoints, and small human scales~\cite{nguyen2022state}.
Early UAV-oriented approaches often relied on 2D CNN backbones such as ResNet~\cite{He_2016_CVPR} and MobileNet~\cite{howard2017mobilenets} to extract per-frame features, followed by temporal aggregation~\cite{mou2020event,mishra2020drone}.
Two-stream CNNs further modeled appearance and motion cues in parallel~\cite{perera2019drone,perera2020multiviewpoint}.
To better capture spatiotemporal dynamics, 3D CNNs~\cite{sultani2021human,li2021uav} extend convolution across the time dimension and are widely adopted in UAV action benchmarks.
More recent work incorporates attention mechanisms and spectral designs, such as AZTR~\cite{Wang2023AZTRAV} and Fourier-based recognition~\cite{divya2022far,kothandaraman2023frequency}, improving temporal reasoning under limited computation.
Despite these advances, most existing UAV action recognition methods remain fully supervised, relying on costly annotations and often requiring non-trivial training/inference recipes, which limits scalability and robustness to unseen aerial domains.
In contrast, we study self-supervised pretraining for UAV videos to learn transferable representations from unlabeled aerial footage while keeping inference simple and efficient.

\textbf{Object-based Video Representation.}
Object-centric reasoning has emerged as a promising direction for video understanding~\cite{elsayed2022savi,NEURIPS2020_8511df98}.
Methods based on region-of-interest (RoI) features~\cite{Wang_2018_ECCV} or detector-driven feature banks~\cite{lfb2019,9577767} provide explicit entity-level representations, enabling models to attend to meaningful objects rather than processing all pixels uniformly.
Transformer-based frameworks such as ORViT~\cite{orvit2021} and ObjectViViT~\cite{zhou2023can} further encode object tokens or object-to-pixel relations to strengthen entity-centric reasoning for recognition.
However, many object-centric pipelines rely on reliable detections and require inference-time region processing, which can be brittle and costly.
In contrast, \emph{FALCON} leverages object cues only during pretraining and performs end-to-end action recognition from raw RGB at inference.

\textbf{Masked Visual Modeling and UAV-Specific Self-Supervision.}
Masked autoencoding (MAE) has become a central paradigm in self-supervised pretraining.
From early denoising autoencoders~\cite{vincent2008extracting} to Transformer-based methods such as BEiT~\cite{bao2021beit} and VideoMAE~\cite{tong2022videomae}, these approaches reconstruct masked patches to learn semantic representations from unlabeled data.
Later variants such as MAR~\cite{qing2023mar} and VideoMAE V2~\cite{wang2023videomae} improve training efficiency via alternative masking designs, yet their masking policies and reconstruction targets are typically semantics-agnostic.
In UAV footage, standard random or tube masking may entirely remove small human regions, and the learning signal can be dominated by reconstructing large background areas under extreme foreground--background imbalance.
Beyond within-clip reconstruction, recent MAE-style objectives explore cross-time reconstruction to encourage temporal correspondence, e.g., reconstructing a heavily masked future view from another time step~\cite{gupta2023siamese}.
However, directly modeling future content in aerial videos can be dominated by ego-motion and background changes, providing weak supervision for small action-relevant regions.

Recent UAV-specific self-supervised or weakly supervised methods~\cite{mliki2020human,li2021uav,choi2020unsupervised} incorporate motion cues or weak supervision, but do not explicitly address the background-dominated learning signal induced by small targets, and provide limited supervision for anticipatory motion dynamics.
In contrast, \emph{FALCON} couples balanced object-aware visibility with object-centric reconstruction supervision for small targets, and re-targets temporal learning through object-centric dual-horizon future reconstruction that is less sensitive to ego-motion/background changes.
\section{Methodology}
\label{methodology}

We present \textbf{FALCON}, a self-supervised pretraining objective for UAV action recognition that is explicitly \emph{object-aware} and \emph{future-aware}.
We first formalize the UAV-specific challenges that make standard masked autoencoding suboptimal (Sec.~\ref{3:problem_formation}).
We then provide an overview of FALCON (Sec.~\ref{subsec:overview}) and detail (i) object-aware masked reconstruction on the observed segment (Sec.~\ref{subsec:observed_object_aware}) and (ii) object-centric dual-horizon future reconstruction on the future segment (Sec.~\ref{subsec:future_prediction}).

\subsection{Problem Formulation}
\label{3:problem_formation}

Masked autoencoding for UAV videos exhibits two coupled objective mismatches.

\textbf{(1) Spatial background dominance.}
UAV videos are dominated by background tokens, while action-relevant human/object regions occupy only a small fraction of the field of view.
With semantics-agnostic masking and uniform reconstruction, the training signal is largely explained by background appearance, yielding representations that under-emphasize action-carrying regions.

\textbf{(2) Limited supervision for motion evolution.}
Standard masked reconstruction mainly recovers missing content within the observed segment, providing limited pressure to encode how human/object regions evolve over time.
For UAV action recognition, however, discriminative cues often lie in the short- and longer-horizon evolution of object-centric motion, which can be subtle under camera motion and cluttered context.

These observations motivate a pretraining objective that (i) ensures sufficient visibility of action-relevant regions under masking, (ii) allocates reconstruction supervision toward those regions, and (iii) injects anticipatory temporal supervision across multiple future horizons.
Accordingly, we focus on three design questions:
\begin{itemize}
    \item \textbf{Balanced token visibility.}
    How can we design masking so that small human/object regions are not systematically excluded from the visible set, while preserving contextual cues?
    \item \textbf{Object-centric supervision allocation.}
    How can we concentrate reconstruction supervision on action-carrying regions to counter background-dominated optimization?
    \item \textbf{Multi-horizon future dynamics.}
    How can we impose lightweight yet effective supervision for both short- and long-horizon motion evolution in an object-centric manner?
\end{itemize}


\subsection{Method Overview}
\label{subsec:overview}

FALCON performs self-supervised pretraining for UAV action recognition with an asymmetric encoder--decoder.
Given an input clip, we split it into an observed clip and a future clip, and optimize a unified reconstruction objective. The pretrained encoder is then transferred to downstream action recognition.

Concretely, FALCON uses two complementary reconstruction objectives.
For the observed clip, it performs \emph{object-aware masked reconstruction}: objectness cues are used to (i) construct a balanced masking pattern that preserves sufficient human/object evidence and (ii) allocate reconstruction supervision toward action-relevant regions to counter background dominance.
For the future clip, it performs \emph{object-centric dual-horizon future reconstruction}: all future tokens are reconstructed from observed context, with supervision restricted to an object-centric region and separated into short- and long-horizon targets to encourage anticipatory dynamics learning. Importantly, off-the-shelf detections are used only to instantiate pretraining cues; no detector input is required during fine-tuning or inference.
\begin{figure}[tp]
    \centering
    \includegraphics[width=0.7\columnwidth]{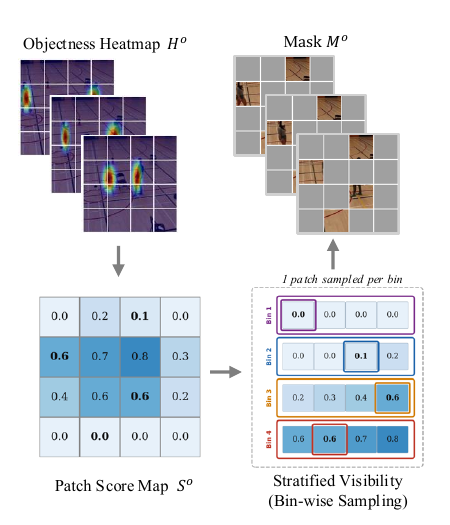}
    \vspace{-10pt}
    \caption{\textbf{Object-aware masking via stratified bin sampling.}
    Detections are aggregated into an objectness heatmap $H^o$ and projected to a patch score map $S^o$.
    Patches are sorted by $S^o$ and partitioned into equal-length bins; we sample one visible patch per bin to form a balanced mask $M^o$, ensuring coverage of both action-relevant (high-score) and contextual (low-score) regions.}
\label{fig:oam}
    \label{fig:masking_Stra} 
\vspace{-15pt}
\end{figure}

\subsection{Object-Aware Reconstruction on Observed Frames}
\label{subsec:observed_object_aware}

Let the input clip be decomposed as $V=\{V^o,V^f\}$, where $V^o\in\mathbb{R}^{T_o\times C\times H\times W}$ denotes the observed clip.
We pretrain on $V^o$ with an object-aware masked reconstruction objective.
Crucially, FALCON couples \emph{token visibility} under masking with \emph{supervision allocation} during reconstruction under a shared objectness prior, which is essential under the extreme foreground--background imbalance in UAV videos.

\paragraph{Objectness prior.}
Given off-the-shelf detections on observed frames, denoted as $\mathbb{B}^o=\{\{b_{t,i}\}_{i=1}^{n_t}\}_{t=1}^{T_o}$, we construct a pixel-level objectness heatmap via Gaussian aggregation:
\begin{equation}
\label{eq:obs_heatmap}
H^o(x,y)=\frac{1}{T_o}\sum_{t=1}^{T_o}\sum_{i=1}^{n_t}
\exp\!\left(
-\frac{(x-x^c_{t,i})^2+(y-y^c_{t,i})^2}{2\sigma^2}
\right),
\end{equation}
where $(x^c_{t,i},y^c_{t,i})$ is the center of $b_{t,i}$.
Projecting $H^o$ to the patch grid yields patch-wise objectness scores
\begin{equation}
\label{eq:obs_patch_objectness}
S^o=\{S_j^o\}_{j=1}^{N_s},\quad N_s=\frac{H}{h}\frac{W}{w}.
\end{equation}

\paragraph{Balanced object-aware masking (stratified visibility).}
To avoid background-dominated visibility, we enforce \emph{stratified} token selection based on $S_j^o$.
Specifically, we sort patches by $S_j^o$, partition them into equal-length bins (objectness quantiles), and select visible tokens to cover the full score range by sampling one visible patch from each bin.
The resulting spatial pattern is replicated across time to form the mask $M^o$.
This design ensures that small human/object regions are not systematically excluded from the encoder input, while still preserving contextual cues.

\paragraph{Object-centric reconstruction objective (supervision allocation).}
Let $K^{inv,o}$ denote the masked observed tokens and let $\hat{V}^o$ be the decoder predictions.
We allocate reconstruction supervision on masked tokens according to objectness:
\begin{align}
\label{eq:loss_obs}
\mathcal{L}_{obs}
&=
\sum_{i\in K^{inv,o}}
\tilde{w}_i^o\,
\|V_i^o-\hat{V}_i^o\|_2^2, \\
\label{eq:obs_weight}
\tilde{w}_i^o
&=
\frac{S_i^o+\mu}{\sum_{j\in K^{inv,o}}(S_j^o+\mu)},
\end{align}
where $\mu$ is the mean objectness score.
The additive $\mu$ provides a background floor to prevent zero-weight tokens and stabilizes optimization, while still prioritizing action-relevant regions.

Overall, this design mitigates UAV background dominance by jointly shaping the \emph{input evidence} (balanced visibility) and the \emph{learning signal} (object-centric supervision allocation) under the same objectness prior.

\subsection{Object-Centric Dual-Horizon Future Reconstruction}
\label{subsec:future_prediction}

Beyond observed-clip reconstruction, FALCON introduces a future-aware objective to explicitly learn motion evolution.
Let $V^f\in\mathbb{R}^{T_f\times C\times H\times W}$ denote the future clip, which we split into two horizons $V^f=\{V^s,V^l\}$ (short/long).
All future tokens are masked and reconstructed from the observed context, while supervision is restricted to an object-centric region to reduce ego-motion/background-dominated learning.

\paragraph{Object-centric supervision region.}
We compute a future objectness heatmap $H^f$ in the same manner as Eq.~\eqref{eq:obs_heatmap}, using pretraining-time detections on future frames.
We then form a high-response support set
\begin{equation}
\label{eq:future_support_set}
\Omega=\{(x,y)\mid H^f(x,y)\ge \gamma\max(H^f)\},
\end{equation}
compute its bounding rectangle, and dilate it by a fixed margin to get the contextual block:
\begin{equation}
\label{eq:future_region}
\mathcal{R}^f=\mathrm{Dilate}(\mathrm{BBox}(\Omega), m),
\end{equation}
where $\mathrm{Dilate}(\cdot,m)$ expands the rectangle by $m$ patch units along each spatial direction (we use $m=1$ throughout all experiments).
Projecting $\mathcal{R}^f$ to the patch grid yields a weight map
\begin{equation}
\label{eq:future_weights}
W^f=\{w_j^f\}_{j=1}^{N_s}.
\end{equation}
In contrast to the token-level objectness scores $S^o$ used on observed clips (Eq.~\eqref{eq:obs_patch_objectness}), $\mathcal{R}^f$ defines a contextual object-motion region that restricts where future reconstruction is supervised, mitigating ego-motion/background-dominated targets; $W^f$ (Eq.~\eqref{eq:future_weights}) is the corresponding region mask/weight map in patch space.

\paragraph{Dual-horizon reconstruction objective.}
Let $K^{inv,s}$ and $K^{inv,l}$ denote masked token sets for short and long horizons, respectively.
Using weights normalized from $W^f$ (optionally with a small background floor), we define two horizon-specific losses:
\begin{align}
\label{eq:loss_short}
\mathcal{L}_{short}
&=
\sum_{i\in K^{inv,s}}
\tilde{w}^{f}_i\,
\|V_i-\hat{V}_i\|_2^2, \\
\label{eq:loss_long}
\mathcal{L}_{long}
&=
\sum_{i\in K^{inv,l}}
\tilde{w}^{f}_i\,
\|V_i-\hat{V}_i\|_2^2,
\end{align}
where $\tilde{w}^{f}_i$ is normalized from $W^f$.

To encourage temporal coherence across horizons, we additionally apply a \emph{consistency} loss:
\begin{equation}
\label{eq:loss_consistency}
\mathcal{L}_{cons}
=
\|\bar{z}^{s}-\bar{z}^{l}\|_2^2,
\end{equation}
where $\bar{z}^{s}$ and $\bar{z}^{l}$ denote the mean decoder-predicted token features over the short- and long-horizon futures, respectively (averaged over spatial tokens within the object-centric region).

\paragraph{Unified objective.}
The full training objective is
\begin{equation}
\label{eq:loss_total}
\mathcal{L}
=
\mathcal{L}_{obs}
+\mathcal{L}_{short}
+\mathcal{L}_{long}
+\mathcal{L}_{cons},
\end{equation}
where $\mathcal{L}_{obs}$ is defined in Eq.~\eqref{eq:loss_obs}. Overall, $\mathcal{L}_{obs}$ provides object-aware masked supervision on observed clips, $\mathcal{L}_{short}$/$\mathcal{L}_{long}$ enforce dual-horizon future reconstruction within the object-centric region, and $\mathcal{L}_{cons}$ encourages coherent predictions across horizons.
\section{Experiments}
\label{experiments}
\begin{table*}[!t]
\centering
\resizebox{0.7\textwidth}{!}{
\begin{tabular}{c c c c c c c c c }
\toprule
Method & Backbone & Extra data  & Input Size & Frames  & GFLOPs & Params. &  \makecell[c]{NEC-Drone\\Acc.@$1$ $\uparrow$ }  & \makecell[c]{UAV-Human\\Acc.@$1$ $\uparrow$ } \\
\midrule 
\multicolumn{9}{c}{\textit{\textbf{supervised}}}\\
FAR~\cite{divya2022far} & X3D-M & K400 & $540 \times 960$ & 8 & 65 & 4M & 71.4 &38.6\\
DiffFAR~\cite{kothandaraman2023frequency} & X3D-M & K400  & $540 \times 960$ & 8  & 130 & 4M & 80.7 &41.9\\
AZTR~\cite{Wang2023AZTRAV} & X3D-M & K400& $224\times224$ & 16  & 7 & 4M & -& 47.4 \\
MITFAS\cite{xian2024mitfas} & X3D-M & K400 & $224\times224$ & 16  & 7& 4M & 78.6 &50.8\\
PMI Sampler~\cite{Xian_PMI}& X3D-M& K400 & $224\times224$ & 16  & 7& 4M & 62.5 & 55.0\\
MViT v1~\cite{Fan_2021_ICCV} & MViT-B& K400 & $224\times224$ & 16  & 71& 37M & 34.6 &24.3\\
ViViT FE~\cite{Arnab2021ViViTAV} & ViT-B& IN-21K & $224\times224$ & 16   & 284& 116M & 38.4&34.1\\
TimesFormer~\cite{gberta_2021_ICML} & ViT-B& K400 & $224\times224$ & 8   & 196& 131M & 40.5&38.4\\
MotionFormer~\cite{patrick2021keeping} & ViT-B& IN21K + K400 & $224\times224$ & 8  & 370 & 109M & 73.6&50.4\\
\midrule
\multicolumn{9}{c}{\textit{\textbf{self-supervised}}}\\
ST-MAE~\cite{feichtenhofer2022masked} & ViT-B & K400 & $224\times224$ & 16  & 180 & 87M & -  & 45.1  \\
VideoMAE~\cite{tong2022videomae} & ViT-B & K400 & $224\times224$ & 16   & 180 & 87M & \underline{82.5}  &\underline{62.1}  \\
MVD~\cite{wang2022masked} & ViT-B & IN21K + K400 & $224\times224$ & 16   & 180 & 87M & 77.6  & 60.5  \\
SiamMAE~\cite{gupta2023siamese} & ViT-B &  K400 & $224\times224$ & 16   & 180 & 87M & 76.4  & 60.1  \\
\textbf{FALCON(Ours)} & ViT-B & K400 & $224\times224$ & 16   & 180 & 87M & \textbf{85.4}&\textbf{67.9}\\
\midrule
VideoMAE~\cite{tong2022videomae} & ViT-L & K400 & $224\times224$ & 16  & 597 & 305M & \underline{88.4}  & \underline{71.5}  \\
VideoMAE v2~\cite{wang2023videomae} & ViT-G  & K400 & $224\times224$ & 16 & 5100 & 632M & 82.2  & 61.1  \\
\textbf{FALCON(Ours)} & ViT-L & K400 & $224\times224$ & 16   & 597 & 305M & \textbf{91.0}&\textbf{77.4}\\

\bottomrule 
\end{tabular}
}
\vspace{-5pt}
\caption{\textbf{Results on NEC-Drone and UAV-Human.}
FALCON sets new state of the art, improving over VideoMAE by +2.9/+5.8 (ViT-B) and +2.6/+5.9 (ViT-L) top-1 on NEC-Drone/UAV-Human.}
\label{tab:necdrone}
\vspace{-15pt}
\label{tab:necdrone}
\end{table*}
\subsection{Datasets}
We evaluate FALCON on two UAV action recognition benchmarks (UAV-Human and NEC-Drone) and two standard ground-view benchmarks (HMDB51 and UCF101). 

\textbf{UAV-Human}~\cite{li2021uav} is a large-scale UAV dataset for human behavior analysis, containing 22,476 videos with 155 action classes captured in diverse indoor and outdoor environments. The aerial viewpoint introduces small targets, dynamic backgrounds, and illumination variations.

\textbf{NEC-Drone}~\cite{choi2020unsupervised} contains 5,250 videos across 16 action classes recorded in a basketball court using a low-altitude UAV. Despite a relatively consistent scene, the data exhibits motion blur and light-reflection noise.

\textbf{HMDB51}~\cite{6126543} consists of 6,766 video clips from 51 action classes. We follow the standard three-split protocol and report average top-1 accuracy.

\textbf{UCF101}~\cite{soomro2012ucf101} contains 13,320 videos from 101 action classes. We follow the standard three-split protocol and report average top-1 accuracy.

We follow the official train/test split for UAV-Human and NEC-Drone, and the standard three-split protocols for HMDB51 and UCF101, reporting average top-1 accuracy.
\begin{table}[!t]
\centering
\resizebox{0.95\columnwidth}{!}{%
\begin{tabular}{l c c c c c}
\toprule
Method & Backbone & Extra data & Param (M) &
\makecell[c]{UCF101\\Acc.@1 $\uparrow$} &
\makecell[c]{HMDB51\\Acc.@1 $\uparrow$} \\
\midrule
Vi$^2$CLR~\cite{diba2021vi2clr} & S3D & K400 & 9 & 89.1 & 55.7 \\
CORP~\cite{hu2021contrast} & Slow-R50 & K400 & 32 & 93.5 & 68.0 \\
CVRL~\cite{qian2021spatiotemporal} & Slow-R50 & K400 & 32 & 92.9 & 67.9 \\
CVRL~\cite{qian2021spatiotemporal} & Slow-R152 & K600 & 328 & 94.4 & 70.6 \\
$\rho$BYOL~\cite{feichtenhofer2021large} & Slow-R50 & K400 & 32 & 94.2 & 72.1 \\
VIMPAC~\cite{tan2021vimpac} & ViT-L & HowTo100M & 307 & 92.7 & 65.9 \\
MVD~\cite{wang2022masked} & ViT-B & IN-1K + K400 & 87 & \underline{97.0} & \underline{76.4} \\
VideoMAE~\cite{tong2022videomae} & ViT-B & K400 & 87 & 96.1 & 73.3 \\
\textbf{FALCON (Ours)} & ViT-B & K400 & 87 & \textbf{97.0} & \textbf{76.6} \\
\bottomrule
\end{tabular}%
}
\vspace{-5pt}
\caption{\textbf{Results on UCF101 and HMDB51.} FALCON improves over VideoMAE by +0.9 on UCF101 and +3.3 on HMDB51.}
\label{tab:ucfhmdb}
\vspace{-10pt}
\end{table}
\subsection{Implementation Details}
We adopt a ViT-based masked autoencoding architecture with either a 12-layer ViT-Base or a 24-layer ViT-Large encoder and an 8-layer narrow ViT decoder, initialized from Kinetics-400 pretrained weights.
Unless otherwise specified, we pretrain for 400 epochs on 16-frame clips at $224\times224$ resolution and then fine-tune the encoder for 100 epochs on downstream action recognition.
We use dense temporal sampling during fine-tuning and evaluation and apply a uniform inference protocol across methods: 5 clips $\times$ 3 crops per video.
\rev{During object-aware pretraining, we use off-the-shelf detections to instantiate object cues: for UAV datasets we focus on human detections to reduce semantic noise from distant background elements, while for HMDB51/UCF101 we additionally include task-relevant object categories (e.g., sports equipment or instruments) to capture contextual cues.}

\subsection{Main Results and Analysis}
Best results are shown in \textbf{bold} and second-best are \underline{underlined} in all the tables. For ViT-based self-supervised baselines that we compare against, we fine-tune and evaluate models using a consistent recipe to ensure a fair comparison.

\textbf{Comparison with State-of-the-Art on the UAV datasets.}
Table~\ref{tab:necdrone} summarizes results on NEC-Drone and UAV-Human.
With the same ViT backbone and Kinetics-400 initialization, FALCON consistently improves over the strongest self-supervised baseline VideoMAE (e.g., +2.9/+5.8 with ViT-B on NEC-Drone/UAV-Human), indicating that incorporating object cues during pretraining yields more action-aligned representations in aerial videos.
FALCON also surpasses the best supervised UAV-specific methods on both benchmarks (e.g., +4.7 on NEC-Drone over DiffFAR and +12.9 on UAV-Human over PMI Sampler with ViT-B), despite not requiring dense action labels during pretraining.
These gains suggest that the key bottleneck in UAV recognition is objective alignment: background-dominated reconstruction and ego-motion/background changes can overwhelm small human-centric cues.
By explicitly centering both masked modeling and future supervision on action-relevant regions, FALCON mitigates this bias and improves robustness to aerial nuisance factors (small targets, clutter, and camera motion).
Notably, simply scaling backbone size or compute budget does not guarantee better UAV performance (e.g., VideoMAE v2 with ViT-G), reinforcing the importance of region-focused objectives for UAV videos.

\textbf{Comparison with State-of-the-Art on the standard datasets.}
Table~\ref{tab:ucfhmdb} reports transfer results on UCF101 and HMDB51.
FALCON consistently improves over the MAE-style baseline VideoMAE (+0.9 on UCF101 and +3.3 on HMDB51), indicating that the proposed object-aware pretraining objective is not restricted to aerial videos.
Notably, FALCON matches the best ViT-B accuracy on UCF101 (tied with MVD) and slightly exceeds it on HMDB51, suggesting that emphasizing action-relevant regions can benefit transfer even when foreground objects are larger and the foreground--background imbalance is less severe than in UAV data.

\begin{table}[tb]
\centering
\resizebox{\columnwidth}{!}{%
\begin{tabular}{l c c c c}
\toprule
Method & Backbone & Extra data &
\makecell[c]{NEC-Drone\\$\rightarrow$ UAV-Human\\Acc.@$1$ $\uparrow$} &
\makecell[c]{UAV-Human\\$\rightarrow$ NEC-Drone\\Acc.@$1$ $\uparrow$} \\
\midrule
VideoMAE~\cite{tong2022videomae} & ViT-B & K400 & 59.4 & 76.6 \\
MVD~\cite{wang2022masked}        & ViT-B & IN-1K + K400 & 58.7 & 76.1 \\
SiamMAE~\cite{gupta2023siamese}        & ViT-B &  K400 & 57.2 & 75.2 \\
\textbf{FALCON (Ours)}           & ViT-B & K400 & \textbf{64.1} & \textbf{80.6} \\
\midrule
VideoMAE~\cite{tong2022videomae} & ViT-L & K400 & 61.6 & 79.1 \\
\textbf{FALCON (Ours)}           & ViT-L & K400 & \textbf{66.0} & \textbf{82.7} \\
\bottomrule
\end{tabular}%
}
\vspace{-4pt}
\caption{\textbf{Transfer Learning Ability.} Cross-dataset transfer between NEC-Drone and UAV-Human.}
\label{tab:transfer}
\vspace{-8pt}
\end{table}
\textbf{Cross-dataset transfer on UAV benchmarks.}
Table~\ref{tab:transfer} evaluates cross-dataset transfer between NEC-Drone and UAV-Human.
FALCON consistently achieves the best transfer accuracy in both directions, improving over VideoMAE by +4.7 (NEC$\rightarrow$UAV) and +4.0 (UAV$\rightarrow$NEC) with ViT-B. Similar improvements hold for ViT-L (+4.4 and +3.6 in the two transfer directions).
These gains indicate stronger robustness to dataset-specific biases and suggest that centering pretraining supervision on action-relevant regions improves generalization across aerial domains. 

\begin{figure*}[tp]
    \centering
    \includegraphics[width=0.85\textwidth]{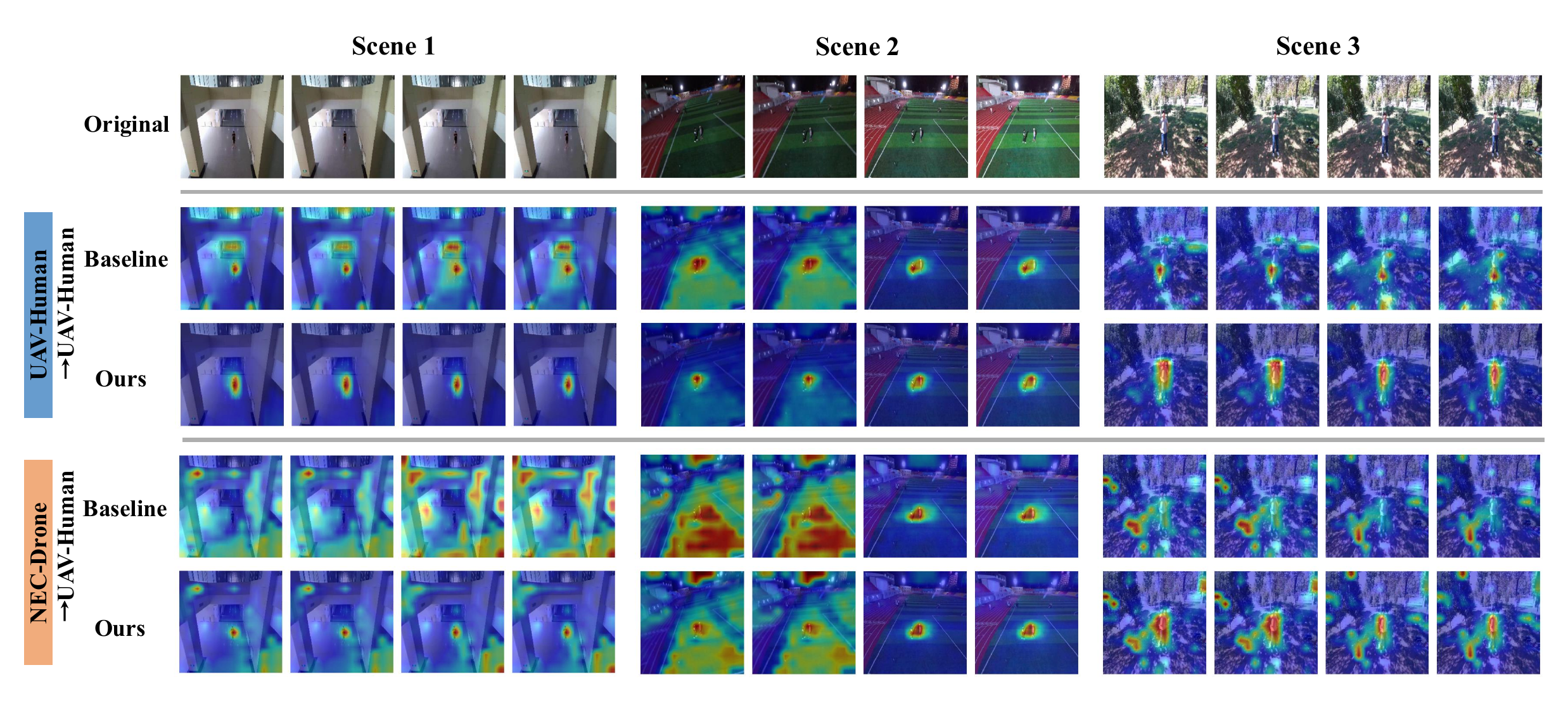}
    \vspace{-10pt}
\caption{\textbf{Attention visualization under in-domain and cross-domain transfer.}
We visualize attention maps on three UAV-Human scenes under two settings: \textbf{in-domain} pretraining on UAV-Human followed by fine-tuning on UAV-Human, and \textbf{cross-domain} pretraining on NEC-Drone followed by fine-tuning on UAV-Human.
Compared to the baseline, FALCON produces more concentrated attention on human regions and suppresses background responses, with a larger improvement under cross-domain transfer.}

\label{fig:attn}
    \vspace{-15pt}
\end{figure*}
\textbf{Attention visualization.}
Fig.~\ref{fig:attn} visualizes attention maps on UAV-Human scenes under both in-domain and cross-domain transfer.
Compared to the baseline, FALCON consistently concentrates attention on the human regions and reduces background activations.
The difference is more pronounced in the cross-domain setting (NEC-Drone$\rightarrow$UAV-Human), qualitatively supporting FALCON's improved transfer robustness in Table~\ref{tab:transfer}.

\begin{table}[!t]
\centering
\resizebox{0.95\columnwidth}{!}{
 \begin{tabular}{c c c c c c }
                \toprule
        Method & Data Aug. & Backbone & \makecell[c]{NEC Drone \\ Acc.@$1$ $\uparrow$} & \makecell[c]{UAV-Human \\ Acc.@$1$ $\uparrow$} & \makecell[c]{Inference time\\ /video (ms)}\\
                 \midrule
        AZTR~\cite{Wang2023AZTRAV} & \cmark & X3D-M& - & 47.4 & \underline{37.1}\\
        MITFAS~\cite{xian2024mitfas} & \cmark& X3D-M & \underline{78.6} &\underline{50.8} &92.4\\
        \textbf{FALCON(Ours)} & \xmark &ViT-B & \textbf{85.4} & \textbf{67.9} & \textbf{18.7}\\
        \bottomrule 
        \end{tabular}
}
\vspace{-5pt}
\caption{\textbf{Inference time comparison.}
On an RTX A5000 GPU, FALCON runs at 18.7\,ms/video, achieving $\sim$2$\times$ and $\sim$5$\times$ speedups over AZTR and MITFAS while improving accuracy.}
\label{tab:inference}
\vspace{-15pt}
\end{table}
\textbf{Inference Time.}
Table~\ref{tab:inference} compares inference latency on an RTX A5000 GPU.
FALCON achieves the lowest per-video latency (18.7 ms/video), yielding $\sim$2$\times$ speedup over AZTR and $\sim$5$\times$ over MITFAS while also improving accuracy on UAV benchmarks.
This efficiency stems from using object cues only during pretraining: at inference, FALCON runs end-to-end on raw RGB clips without online detection or additional alignment modules required by prior UAV-specific pipelines. All timings measure end-to-end forward inference per video under the same hardware and input settings; no test-time augmentation is used for FALCON.

\begin{table}[tb]
\centering

\resizebox{\columnwidth}{!}{%
\begin{tabular}{l c c c c c c}
\toprule
Model Variant
& $\mathcal{L}_\text{obs}$ & OAM
& $\mathcal{L}_\text{short}$ & $\mathcal{L}_\text{long}$ & $\mathcal{L}_\text{cons}$
& Acc.@$1$ $\uparrow$ \\
\midrule
Baseline                           &            &            &            &            &            & 62.1 \\
\quad + OAM                        &            & \checkmark &            &            &            & 64.0 \\
\quad + $\mathcal{L}_\text{obs}$   & \checkmark & \checkmark &            &            &            & 66.4 \\
\quad + $\mathcal{L}_\text{short}$ & \checkmark & \checkmark & \checkmark &            &            & 66.8 \\
\quad + $\mathcal{L}_\text{long}$  & \checkmark & \checkmark & \checkmark & \checkmark &            & 67.1 \\
\textbf{Full FALCON}               & \checkmark & \checkmark & \checkmark & \checkmark & \checkmark & \textbf{67.9} \\
\bottomrule
\end{tabular}%
}
\vspace{-4pt}
\subcaption{\textbf{(a) Core Component Ablation.} Each component independently contributes.}
\label{tab:component}
\vspace{-10pt}
\bigskip

\begin{minipage}[t]{0.49\columnwidth}
\centering
\resizebox{\columnwidth}{!}{%
\begin{tabular}{l c}
\toprule
Future Design & Acc.@$1$ $\uparrow$ \\
\midrule
No future prediction             & 66.4 \\
Single horizon (uniform)         & 67.1 \\
Short-horizon only               & 66.8 \\
Long-horizon only                & 66.5 \\
Bidirectional (past + future)    & 67.0 \\
\textbf{Dual-horizon (ours)}     & \textbf{67.9} \\
\bottomrule
\end{tabular}%
}
\vspace{-4pt}
\subcaption{\textbf{(b) Future Horizon Design.}}
\label{tab:horizon}
\end{minipage}
\hfill
\begin{minipage}[t]{0.49\columnwidth}
\centering
\resizebox{\columnwidth}{!}{%
\begin{tabular}{l c}
\toprule
Supervision Region & Acc.@$1$ $\uparrow$ \\
\midrule
Full frame                 & 66.1 \\
Tight bounding box         & 67.5 \\
Heatmap weights        & 66.5 \\
\textbf{Expanded $\mathcal{R}^f$ (ours)} & \textbf{67.9} \\
\bottomrule
\end{tabular}%
}
\vspace{-4pt}
\subcaption{\textbf{(c) Future Supervision Region.}}
\label{tab:block}
\end{minipage}
\vspace{-10pt}
\bigskip

\begin{minipage}[t]{0.49\columnwidth}
\centering
\resizebox{\columnwidth}{!}{%
\begin{tabular}{l c c}
\toprule
\multirow{2}{*}{Masking} & \multicolumn{2}{c}{Acc.@$1$ $\uparrow$} \\
\cmidrule(lr){2-3}
 & w/o Future & w/ Future \\
\midrule
Random                       & 63.7 & 65.5 \\
Tube~\cite{tong2022videomae}  & 64.1 & 66.3 \\
Block                        & 62.1 & 61.8 \\
\textbf{Ours}                & \textbf{66.4} & \textbf{67.9} \\
\bottomrule
\end{tabular}%
}
\vspace{-4pt}
\subcaption{\textbf{(d) Masking Strategy.}}
\label{tab:masking}
\end{minipage}
\hfill
\begin{minipage}[t]{0.49\columnwidth}
\centering
\resizebox{\columnwidth}{!}{%
\begin{tabular}{c c c}
\toprule
\#Observed & \#Future & Acc.@$1$ $\uparrow$ \\
\midrule
12 & 4 & \textbf{67.9} \\
10 & 6 & 67.5 \\
8  & 8 & 67.2 \\
\bottomrule
\end{tabular}%
}
\vspace{-4pt}
\subcaption{\textbf{(f) Observed vs.\ Future Split.}}
\label{tab:split}
\end{minipage}
\vspace{-10pt}
\bigskip

\begin{minipage}[t]{0.54\columnwidth}
\centering
\resizebox{\columnwidth}{!}{%
\begin{tabular}{c c c c c c c}
\toprule
Mask Ratio & 50\% & 60\% & 70\% & 80\% & 90\% & 95\% \\
\midrule
Acc.@$1$ $\uparrow$ & 67.8 & 67.9 & \textbf{67.9} & 67.3 & 66.2 & 64.7 \\
\bottomrule
\end{tabular}%
}
\vspace{-4pt}
\subcaption{\textbf{(e) Masking Ratio.}}
\label{tab:maskratio}
\end{minipage}
\hfill
\begin{minipage}[t]{0.42\columnwidth}
\centering
\resizebox{\columnwidth}{!}{%
\begin{tabular}{c c c c c}
\toprule
Videos with boxes & 25\% & 50\% & 75\% & 100\% \\
\midrule
Acc.@$1$ $\uparrow$ & 61.8 & 65.4 & 67.1 & \textbf{67.9} \\
\bottomrule
\end{tabular}%
}
\vspace{-4pt}
\subcaption{\textbf{(g) Partial Bboxes.}}
\label{tab:bbox}
\end{minipage}
\vspace{-10pt}
\bigskip

\resizebox{\columnwidth}{!}{%
\begin{tabular}{l l c c}
\toprule
Detector & Backbone
& \makecell[c]{CrowdHuman \\ mAP $\uparrow$}
& \makecell[c]{UAVHuman \\ Acc.@$1$ $\uparrow$} \\
\midrule
None (Vanilla VideoMAE)     & --        & --   & 62.1 \\
Cascade Mask R-CNN          & MobileNet & 65.7 & 63.9 \\
Faster R-CNN                & HRNet     & 72.3 & 66.1 \\
\textbf{Cascade Mask R-CNN} & HRNet     & 84.1 & \textbf{67.9} \\
\bottomrule
\end{tabular}%
}
\vspace{-4pt}
\subcaption{\textbf{(h) Detector Quality.} Consistent gains even with weaker detections.}
\label{tab:detector}

\caption{\textbf{Ablation Studies.} OAM: object-aware masking. Results on UAVHuman unless noted.}
\vspace{-15pt}
\label{tab:ablation}
\end{table}
\subsection{Ablation Studies}
\label{subsec:ablation}

We conduct ablations on UAV-Human and summarize results in Table~\ref{tab:ablation}.
Unless otherwise stated, all ablations use ViT-B and the same training recipe as the main experiments.

\noindent\textbf{Ablation setup and terminology.}
In Table~\ref{tab:block}, \emph{Full frame} supervises future reconstruction on all spatial tokens; \emph{Tight bounding box} restricts supervision to the union box of detected humans; \emph{Heatmap pixel-level} applies the detection-derived heatmap directly as pixel/patch weights without forming a contextual region; and \emph{Expanded $\mathcal{R}^f$} uses the dilated object-motion region defined in Eq.~\eqref{eq:future_region}.
In Table~\ref{tab:bbox}, \emph{BBox ratio} denotes the fraction of training videos for which detection boxes are available (the remaining videos use no boxes), simulating partial access to object cues.

\textbf{Core components.}
Table~\ref{tab:component} progressively adds each part of FALCON to a VideoMAE-style baseline.
Object-aware masking (OAM) already improves performance, and introducing the observed reconstruction objective $\mathcal{L}_{obs}$ further strengthens learning by jointly shaping token visibility and supervision allocation toward action-relevant regions.
Adding future objectives yields additional gains: $\mathcal{L}_{short}$ and $\mathcal{L}_{long}$ provide complementary supervision for short- and long-horizon dynamics, while the horizon-consistency term $\mathcal{L}_{cons}$ gives a final improvement, indicating that each term contributes and the full objective is most effective.

\textbf{Temporal design: horizon, supervision region, and observed/future split.}
Table~\ref{tab:horizon} compares alternative future designs and shows that the proposed dual-horizon objective outperforms single-horizon variants, suggesting that separating short- and long-range evolution provides a stronger temporal signal than a uniform horizon.
Table~\ref{tab:block} studies \emph{where} future reconstruction is supervised: full-frame supervision is less effective in UAV videos where ego-motion and background changes can dominate the target, while restricting supervision to object-centric regions improves results.
The expanded contextual region $\mathcal{R}^f$ performs best, supporting the robustness of block-level context for small targets and viewpoint jitter.
Table~\ref{tab:split} further examines the observed/future split and indicates that allocating more frames to the observed context is beneficial, while retaining a future window for temporal supervision.

\textbf{Spatial masking choices.}
Table~\ref{tab:masking} compares masking strategies with and without future supervision.
Standard random or tube masking is consistently weaker than the proposed OAM, and block masking is particularly unfavorable in UAV videos where small human regions can be easily removed.
Table~\ref{tab:maskratio} shows that performance is relatively stable over a broad range of masking ratios, while extremely high masking degrades accuracy as expected.

\textbf{Robustness to object cues.}
Table~\ref{tab:bbox} demonstrates that FALCON benefits from stronger object cues: performance improves monotonically as bounding-box availability increases, yet the method remains functional when boxes are only partially available.
Finally, Table~\ref{tab:detector} evaluates detector quality; stronger detectors yield better accuracy, but gains remain consistent even with weaker detections, indicating that FALCON does not hinge on perfect boxes and that detection cues mainly serve as lightweight pretraining priors.
\section{Conclusion, Limitations, and Future Work}

\noindent We presented \textbf{FALCON}, a self-supervised pretraining objective for UAV action recognition that re-targets both masked modeling and temporal supervision to action-relevant human regions.
FALCON couples balanced object-aware visibility with object-centric supervision allocation on observed clips, and introduces object-centric dual-horizon future reconstruction to encourage anticipatory motion modeling.
Across UAV benchmarks, FALCON achieves state-of-the-art accuracy while keeping inference end-to-end from raw RGB clips without detector inputs.FALCON relies on off-the-shelf detections during pretraining to instantiate object cues; improving robustness to noisy detections and reducing this dependency are important directions.
Our current instantiation is evaluated with ViT-based encoders and primarily focuses on human-centric cues in UAV videos.
Future work includes extending the objective to lightweight or hybrid backbones, strengthening longer-horizon temporal modeling.


\bibliographystyle{IEEEtran}
\bibliography{root}
\
\end{document}